\documentclass[11pt]{article}

\usepackage[final]{acl}

\usepackage{times}
\usepackage{latexsym}

\usepackage[T1]{fontenc}

\usepackage[utf8]{inputenc}

\usepackage{microtype}

\usepackage{inconsolata}

\usepackage{inconsolata}
\usepackage{array}
\usepackage{booktabs}
\usepackage{multirow}
\usepackage{float}
\usepackage{graphicx}
\usepackage[dvipsnames]{xcolor}
\usepackage[most]{tcolorbox}
\usepackage{soul}
\usepackage{tikz}
\usetikzlibrary{calc}
\usetikzlibrary{decorations.pathmorphing}

\makeatletter

\newcommand{\defhighlighter}[3][]{%
  \tikzset{every highlighter/.style={color=#2, fill opacity=#3, #1}}%
}

\defhighlighter{yellow}{.5}

\newcommand{\highlight@DoHighlight}{
  \fill [ decoration = {random steps, amplitude=1pt, segment length=15pt}
        , outer sep = -15pt, inner sep = 0pt, decorate
        , every highlighter, this highlighter ]
        ($(begin highlight)+(0,8pt)$) rectangle ($(end highlight)+(0,-3pt)$) ;
}

\newcommand{\highlight@BeginHighlight}{
  \coordinate (begin highlight) at (0,0) ;
}

\newcommand{\highlight@EndHighlight}{
  \coordinate (end highlight) at (0,0) ;
}

\newdimen\highlight@previous
\newdimen\highlight@current

\DeclareRobustCommand*\highlight[1][]{%
  \tikzset{this highlighter/.style={#1}}%
  \SOUL@setup
  \def\SOUL@preamble{%
    \begin{tikzpicture}[overlay, remember picture]
      \highlight@BeginHighlight
      \highlight@EndHighlight
    \end{tikzpicture}%
  }%
  \def\SOUL@postamble{%
    \begin{tikzpicture}[overlay, remember picture]
      \highlight@EndHighlight
      \highlight@DoHighlight
    \end{tikzpicture}%
  }%
  \def\SOUL@everyhyphen{%
    \discretionary{%
      \SOUL@setkern\SOUL@hyphkern
      \SOUL@sethyphenchar
      \tikz[overlay, remember picture] \highlight@EndHighlight ;%
    }{%
    }{%
      \SOUL@setkern\SOUL@charkern
    }%
  }%
  \def\SOUL@everyexhyphen##1{%
    \SOUL@setkern\SOUL@hyphkern
    \hbox{##1}%
    \discretionary{%
      \tikz[overlay, remember picture] \highlight@EndHighlight ;%
    }{%
    }{%
      \SOUL@setkern\SOUL@charkern
    }%
  }%
  \def\SOUL@everysyllable{%
    \begin{tikzpicture}[overlay, remember picture]
      \path let \p0 = (begin highlight), \p1 = (0,0) in \pgfextra
        \global\highlight@previous=\y0
        \global\highlight@current =\y1
      \endpgfextra (0,0) ;
      \ifdim\highlight@current < \highlight@previous
        \highlight@DoHighlight
        \highlight@BeginHighlight
      \fi
    \end{tikzpicture}%
    \the\SOUL@syllable
    \tikz[overlay, remember picture] \highlight@EndHighlight ;%
  }%
  \SOUL@
}
\makeatother

\title{Compact Prompting in Instruction-tuned LLMs for Joint Argumentative Component Detection}

\author{
\textbf{Sofiane Elguendouze}\textsuperscript{1} \quad
\textbf{Erwan Hain}\textsuperscript{1} \quad
\textbf{Elena Cabrio}\textsuperscript{1} \quad
\textbf{Serena Villata}\textsuperscript{1}\\
\textsuperscript{1}Université Côte d’Azur, I3S, CNRS, Inria (Marianne), France\\
\texttt{\{name.lastname\}@univ-cotedazur.fr}
}

\begin{document}
\maketitle
\begin{abstract}
Argumentative component detection (ACD) is a core subtask of Argument(ation) Mining (AM) and one of its most challenging aspects, as it requires jointly delimiting argumentative spans and classifying them into components such as claims and premises. While research on this subtask remains relatively limited compared to other AM tasks, most existing approaches formulate it as a simplified sequence labeling problem, component classification, or a pipeline of component segmentation followed by classification. In this paper, we propose a novel approach based on instruction-tuned Large Language Models (LLMs) using compact instruction-based prompts, and reframe ACD as a language generation task, enabling arguments to be identified directly from plain text without relying on pre-segmented components. Experiments on standard benchmarks show that our approach achieves higher performance compared to state-of-the-art systems. To the best of our knowledge, this is one of the first attempts to fully model ACD as a generative task, highlighting the potential of instruction tuning for complex AM problems. Our code and the datasets used are openly available in the following \href{https://anonymous.4open.science/r/acd-F342/README.md}{GitHub repository}.
\end{abstract}

\section{Introduction}
\label{sec:intro}
Argument(ation) Mining (AM) \cite{10.1145/2850417, cabrio2018five, Stede2019, 10.1162/coli_a_00364} has emerged as a prominent research area within natural language processing, aiming to automatically identify and analyze the structure of arguments expressed in text. By decomposing argumentative discourse into meaningful components and relations, AM supports a wide range of downstream applications, including opinion analysis in large-scale deliberation, decision support, and educational technologies. Among the various subtasks in AM, Argumentative Component Detection (ACD), which consists of the identification and classification of argumentative components such as claims and premises, remains one of the most challenging. This difficulty stems from the need to simultaneously delimit component boundaries and assign correct labels, often in the presence of implicit reasoning and loosely structured discourse. Most existing approaches to ACD continue to frame the task as a sequence labelling problem often relying on extensive feature engineering \cite{stab2014identifying}, multi-stage pipelines (component delimitation + component classification) \cite{morio-etal-2022-end} or a mere component type classification that depends on the availability of pre-segmented argumentative components \cite{chen-etal-2024-exploring-potential, cabessa-etal-2025-argument}. 

Recent advances in large language models (LLMs), particularly instruction-tuned models, have significantly reshaped the landscape of many NLP tasks. Their strong generative capabilities and sensitivity to task descriptions offer new opportunities for addressing complex prediction problems beyond traditional classification paradigms. It is therefore unsurprising that LLM-based approaches have recently been explored in AM \cite{gorur-etal-2025-large, cabessa-etal-2025-argument, favero2025leveragingsmallllmsargument}, primarily focusing on argument classification and argument relation extraction.

Despite leveraging large pre-trained models, ACD is still predominantly formulated as a sequence classification problem operating on pre-segmented (or pre-identified) argumentative units \cite{10.1007/978-3-031-70816-9_20, cabessa-etal-2025-argument}. While such formulations can yield competitive performance, they inherently assume the availability of accurately segmented argumentative spans, reducing the problem to component classification rather than jointly addressing boundary detection and labelling. This assumption is itself highly non-trivial as it constrains applicability in realistic settings where argumentative boundaries are not explicitly/easily marked.

In this work, we propose a novel perspective on ACD by reimagining it as a language generation task. Specifically, we introduce a compact prompting approach for instruction-tuned LLMs, in which models are fine-tuned using concise, instruction-based prompts designed to guide both the delimitation (segmentation) and classification of argumentative components in a joint manner. We evaluate our approach on standard AM benchmarks and show that it achieves performance that surpasses state-of-the-art methods. Importantly, beyond competitive results, this work represents one of the early attempts to reconceptualize ACD as a generative task within an instruction-tuning paradigm. By modeling segmentation and classification jointly as a single text-to-structure generation process, we depart from conventional multi-stage pipelines that decompose the task into separate boundary detection and labeling steps. Our findings suggest that treating ACD as a unified generation problem not only simplifies the modeling framework but also offers a flexible and effective alternative to traditional formulations of AM subtasks, opening new directions for future research in the field.

\section{Related Work}

\subsection{Argument Mining Models}
A wide range of methods has been introduced for AM, initially exploiting manually designed syntactic and lexical features \cite{stab2014identifying}.
Subsequent work employed traditional supervised machine learning algorithms, such as 
Support Vector Machines \cite{stab2017, habernal-gurevych-2017-argumentation}. \cite{Menini_Cabrio_Tonelli_Villata_2018} proposed multi-class Support Vector Machine classifiers which achieved good results on the relation classification stage of the AM pipeline. One of the early learning-based approaches consisted of using more advanced neural network-based models including RNNs \cite{niculae-etal-2017-argument, potash-etal-2017-heres}.


Further studies predominantly adopted supervised learning paradigms based on auto-encoder transformer architectures such as BERT \cite{devlin-etal-2019-bert}, motivated by their strong ability to capture contextual information and long-range argumentative dependencies. For instance, \cite{goffredo:hal-03873412} tackled the task of fallacious argument classification in political debates, specifically those of the U.S. Presidential Campaigns. They first built a large corpus of political debates annotated with fallacious arguments. Then, they defined a transformer-based model architecture for fallacy classification, fine-tuned on argumentation features outperforming standard baselines. \cite{morio-etal-2022-end} introduced a multi-task training framework (MT-AM) that jointly learns from auxiliary corpora to improve performance across AM tasks. They designed an end-to-end model using a two-staged method (multi-task pre-training and target corpus fine-tuning). The approach allowed to integrate information across datasets by sharing a unified model backbone while maintaining corpus-specific output layers. Their system demonstrated that cross-corpus transfer can boost results, particularly for smaller target datasets. \cite{habernal2024mining} investigated AM in decisions of the European Court of Human Rights (ECHR), developing models built on pre-trained BERT and RoBERTa \cite{liu2019roberta}.

LLMs have become the dominant tools in NLP, showing competitive performance across multiple tasks. \cite{chen-etal-2024-exploring-potential} explored LLMs by focusing on “counter speech generation”. They assessed the performance of multiple LLMs on a broad set of computational argumentation tasks (argument mining and argument generation), under zero-shot and few-shot settings. Their findings demonstrated that LLMs surpassed baseline models on certain datasets. \cite{pojoni2023argument} proposed a method that first transcribes podcast episodes into text and then uses OpenAI’s GPT-4 (via ChatGPT) to extract argumentative components (main claim, premise, counterargument and rebuttal). The definition of an argument unit was however different from that in the literature, where it was considered as a statement synthesized from a direct quotation from the text and not an extracted portion of the text. 

\subsection{Argumentative Component Detection}
\label{sec:acd}
\cite{goudas_2014} proposed a multi-stage framework for ACD in social media texts, first classifying sentences as argumentative or non-argumentative, and then applying a Conditional Random Field (CRF) model to identify the exact claim and premise spans within argumentative sentences.
\cite{stab2017} modeled ACD as a sequence labeling task and established strong baselines for identifying claims and premises in noisy online discussions including CRF and feature-rich models combining lexical, syntactic, discourse, and embedding features.
\cite{favero2025leveragingsmallllmsargument} applied small open-source LLMs (Qwen 2.5 7B, Llama 3.1 8B, and Gemma 2 9B) with few-shot prompting and fine-tuning for argument segmentation and classification as a multi-stage pipeline. Their experiments were conducted on a single corpus, PERSUADE 2.0, which consists of argumentative essays written by English high-school students. They shown that fine-tuned models consistently outperform few-shot prompting across both tasks highlighting the benefits of task-specific adaptation even for smaller LLMs. 
\cite{chen-etal-2024-exploring-potential} evaluated LLMs such as GPT 3.5, Flan T5, and LLaMA 2 in zero-shot and few-shot settings across multiple AM tasks including claim and evidence classification (treated separately), stance detection, argument generation, argument summarization, and counter-speech generation. Their results demonstrate that LLMs achieve promising performance across these tasks. However, as illustrated in the prompt formulations provided in their study, the tasks related to argument detection are operationalized primarily as classification problems rather than as extraction tasks where argumentative units are assumed to be pre-segmented. Their approach remains dependent on the prior availability of these segmented argumentative units and sidestep the more challenging problem of jointly identifying and delimiting argumentative components directly from raw text. The work by \cite{cabessa-etal-2025-argument} also explored several LLMs reporting state-of-the-art performance across multiple AM benchmarks. Their study demonstrated effectiveness for tasks such as argumentative component classification, argument relation identification, and relation type classification. However, their focus remains limited to classification-based formulations as previously mentioned without argument boundary detection.

\section{Methodology}
\subsection{Task formalization}
As outlined in Section \ref{sec:intro}, we focus exclusively on the task of ACD (Argumentative Component Detection), one of the core challenges in AM. Unlike joint end-to-end frameworks that simultaneously address component detection and relation prediction \cite{eger2017neural}, we deliberately narrow our scope to ACD as recent state-of-the-art results indicate that argument relation classification has reached relatively high levels of effectiveness. The accurate identification and delimitation of argumentative components from raw text, in contrast, remains insufficiently explored and continues to present substantial challenges. In this line, we aim to jointly perform argumentative unit segmentation and component classification within a unified framework. As discussed in Section \ref{sec:acd}, many state-of-the-art approaches reduce ACD to a classification task applied to pre-segmented argumentative units, a formulation that simplifies the problem but assumes the prior availability of accurately segmented spans, something that rarely holds in real-world scenarios. 

Our approach instead seeks to identify argumentative components directly from plain, unsegmented text. We consider two types of components: \textit{claims} and \textit{premises}. A \textit{claim} is a statement asserting a position, opinion, or proposition that can be supported or contested, often in relation to a debatable topic. A \textit{premise} is a supporting or opposing statement that provides justification, evidence, or reasoning for a claim (or another premise), including statistics, expert testimony, factual information, anecdotes, or illustrative examples. For instance, in the example below, the segment highlighted in orange represents a claim, while the segment in blue corresponds to a premise supporting that claim:
{\begin{center}
    `` ... \highlight[red]{the government should try to preserve minority languages}. This is because \highlight[cyan]{language can be seen as much more than just one method of communication} ... "
\end{center}

\subsection{Datasets}
\label{sec:data}
Three datasets have been used in our experiments: 
\textbf{(1) USElecDeb60To16} \cite{ijcai2019p944}: This dataset consists of annotated transcripts from televised U.S. presidential election debates spanning the period 1960–2016. The corpus captures spontaneous, spoken political discourse characterized by interruptions, rhetorical strategies, and implicit argumentation. Its dialogical structure and conversational nature make argumentative component detection particularly challenging, as argumentative boundaries are often less explicit. Furthermore it is considered as the largest argument-annotated dataset to date.
\textbf{(2) Persuasive Essays} \cite{stab2017}: Initially introduced in \cite{stab-gurevych-2014-annotating}, this dataset has been later extended to finally contain 402 English essays collected from \href{essayforum.com}{essayforum.com}, an online platform where users seek feedback on written compositions such as essays and research papers. Compared to USElecDeb60To16, this corpus exhibit clearer argumentative organization and more explicit discourse markers, making it one of the most widely adopted benchmarks for ACD. \textbf{(3) Web Discourse} \cite{habernal-gurevych-2017-argumentation}: This dataset comprises user-generated web texts addressing six controversial education-related topics. Unlike the relatively well-structured persuasive essays, this corpus reflects informal online discourse, including noisy language, heterogeneous writing styles, and loosely structured argumentation. 

Importantly, these datasets vary substantially in their sizes, writing style, structural clarity and discourse complexity. This diversity exposes models to different linguistic phenomena and levels of argumentative explicitness, thereby equipping our approach with a more robust evaluation setting and allowing us to assess its generalization ability across textual genres with variable argumentation patterns. Table \ref{tab:datasets} provides statistics on these datasets (before the train/dev/test splitting). Further detailed statistics can be found in the appendix in Table \ref{tab:comp_data_detailed}.

\begin{table}[ht]
\centering
\resizebox{.485\textwidth}{!}{%
\begin{tabular}{|c|cc|}
\hline
    Dataset & \multicolumn{2}{l|}{Argument Components} \\ \cline{2-3}
    \multirow{2}{*}{} & \multicolumn{1}{c|}{Claim}    & Premise \\ \hline
    USElecDeb60To16 \cite{ijcai2019p944}& \multicolumn{1}{c|}{29k} & 26k \\ \hline
    Persuasive Essays\cite{stab2017}& \multicolumn{1}{c|}{2257} & 3832 \\ \hline
    Web Discourse \cite{habernal-gurevych-2017-argumentation}& \multicolumn{1}{c|}{195} & 538 \\ \hline
    Merge & \multicolumn{1}{c|}{31.5k} & 30.3k \\ \hline
\end{tabular}}
\caption{Datasets statistics}
\label{tab:datasets}
\end{table}

\subsection{Method}
To perform ACD using LLMs, we reconceptualize the argument detection process as a text generation problem. We prompt the LLM to reproduce the original plain text while inserting argumentative tags that explicitly demarcate the boundaries and nature of argumentative components. This is achieved through a standardized prompting scheme. We employ a prompt template that specifies the detailed task description, the input text, and the exact structure and format expected for the output.

Since our approach relies on instruction-tuned LLMs, the prompts are constructed directly from the annotated datasets described above. To align the data with a generative formulation of the task, all corpora originally annotated using the BIO-tagging scheme\footnote{BIO (Beginning–Inside–Outside) is a common labeling scheme in sequence tagging tasks used to indicate the boundaries of structured elements within text.} are converted into XML-tagged text, where argumentative components are explicitly marked using tags such as \texttt{<premise>...</premise>} and \texttt{<claim>...</claim>} (see Figure \ref{fig:eg_prompt_answer} in the appendix). For each instance, we create pairs consisting of the original plain text and its corresponding XML-tagged version. These pairs serve as input–output examples for instruction tuning. A single reference output may contain multiple argumentative components.

\subsection{Models}
\label{sec:am_models}
We fine-tuned a set of open-weight LLM models on the datasets described in Section \ref{sec:data}. The selected models span different architectural families and parameter scales and include: GPT-2-XL-1.5B \cite{radford2019language}, OPT-1.3B and
OPT-6.7B \cite{zhang2022opt}, Mistral-7B-v0.3 \cite{jiang2023mistral7b}, and Llama-3-8B-Instruct \cite{grattafiori2024llama3herdmodels}. Our decision to rely exclusively on open-weight models despite the availability of more recent and powerful proprietary systems such as GPT-4o is motivated by the fact that open models ensure full reproducibility and transparency of experimental results and allow controlled fine-tuning and architectural inspection, which is not possible with closed APIs. Also, they facilitate broader adoption in real-world settings where cost and data privacy constraints may limit the use of proprietary systems. 

We also include encoder-based transformer architectures in our experiments, namely RoBERTa and DeBERTa \cite{he2021debertadecodingenhancedbertdisentangled}, which are widely used in sequence labeling and classification tasks. Unlike the generative LLMs, these models are trained under traditional token-level classification settings for ACD.

\section{Experimental settings and results}
\subsection{Experimental settings}
For the LLM-based experiments, decoding is configured to minimize stochasticity during generation. Specifically, we employ a very low temperature (0.01) combined with a restrictive nucleus sampling parameter (top-p = 0.1), encouraging near-deterministic outputs. To accommodate context window constraints, longer inputs are segmented into chunks of up to 1024 tokens. Each model is fine-tuned for a minimum of 10 epochs, and the checkpoint achieving the highest macro-F1 score on the validation set is selected for final evaluation. All experiments follow a standard 80/10/10 train–validation–test split.
\begin{table*}[ht]
    \centering
    \resizebox{.65\textwidth}{!}{%
    \begin{tabular}{|lcc|}
    \hline
        Model/Dataset & Macro F1 & Acc\\\hline
        Human Upper Bound \cite{stab2017} - PE & \textbf{0.8860} & -\\\hline\hline
        CRF \cite{goudas_2014} &  0.4237 & - \\\hline
        Heuristic Baseline\cite{stab2017} - PE &  0.642 & - \\\hline
        CRF with features \cite{stab2017} - PE & 0.8670 & - \\\hline
        MT-all \cite{morio-etal-2022-end} - PE& 0.7566 & - \\\hline
        OPT-6.7B - PE & 0.8518 & 0.8856 \\\hline
        GPT-2-1.5B - PE & 0.8521 & 0.8804 \\\hline
        Llama-3-8B - PE & \textbf{0.8778} & 0.9004 \\\hline
        DeBERTa-v3 - PE & 0.7112 & -\\\hline
        RoBERTa - PE & 0.6933 & -\\\hline\hline
        GPT-2-1.5B - Merge& 0.7684 & 0.7955 \\\hline
        OPT-1.3B - Merge & 0.7679 & 0.7971 \\\hline
        OPT-6.7B - Merge & \textbf{0.7822} &  0.8076 \\\hline
        Llama-3-8B - Merge & 0.7667 & 0.8132 \\\hline
        Mistral-7B - Merge & 0.7652 & 0.8016 \\\hline
        DeBERTa-v3 - Merge & 0.49 & -\\\hline
        RoBERTa - Merge & 0.48 & -\\\hline
    \end{tabular}}
    \caption{ACD model assessment. \texttt{PE} denotes the Persuasive Essays dataset, \texttt{Merge} is the combination of all three datasets. Rows without references correspond to our proposed methods/models}
    \label{tab:acd-results}
\end{table*}

For the encoder-based baselines, ACD is formulated as a token-level sequence labeling task using the conventional BIO scheme, where models are fine-tuned for token classification over at least 10 epochs with a learning rate of 1e-4 and a maximum input length of 64 tokens.

\subsection{Results and discussion}
Table \ref{tab:acd-results} reports the performance of our models on the ACD task in terms of macro F1 score and overall accuracy across multiple experimental configurations. We focus on two primary setting: 
(1) Models trained and tested on the \texttt{Merged} dataset, which combines all three corpora described in Section \ref{sec:data}. This setting allows us to assess the robustness of our approach across heterogeneous writing styles and discourse complexity as previously outlined in Section \ref{sec:data}. (2) For comparison with established baselines, we additionally report results obtained when training and testing exclusively on the Persuasive Essays (PE) benchmark. These models are denoted with the suffix \texttt{PE} in Table \ref{tab:acd-results}.

The fine-tuned Llama-3-8B model on the \texttt{PE} dataset achieved the highest macro-F1 score (0.8778) among all methods, outperforming every reported baseline, including the strongest feature-engineered CRF model \cite{stab2017}. It is worth highlighting that the latter relies on syntactic parsing and extensive task-specific hand-crafted features, whereas our instruction-tuned models learn directly through generative supervision. Notably, the gap between our best-performing model and the reported human upper bound (0.8860) is only $8.2 \times 10^{-3}$, indicating that the proposed approach approaches near-human agreement levels.

When trained and evaluated on the \texttt{Merge} dataset, all models exhibit a performance drop compared to the \texttt{PE} setting. This decrease (0.0956) can be attributed to the increased heterogeneity of the combined corpus. The broader variety of discourse styles and argumentative patterns introduce greater complexity and makes boundary detection more challenging. Nevertheless, the best-performing model (OPT-6.7B) still achieves a very satisfying score (0.7822) for this task of ACD.

Encoder-based transformer baselines (RoBERTa and DeBERTa-v3) trained on \texttt{PE} achieve good results and surpass the earlier heuristic baseline. However, both models remain below the performance of the LLM models. These models experience a much sharper degradation in the \texttt{Merge} setting (0.49), suggesting that traditional encoder-based models struggle more with capturing cross-domain variability than generative models. 

Finally, The CRF model from \cite{goudas_2014} was only included for historical comparison purposes. It should be noted that their experiments were conducted on a different dataset (Greek social media text), which differs in language, domain, and annotation scheme from the PE corpus. As a result, direct performance comparison should be interpreted cautiously.

The superiority of the generative formulation can be attributed to several factors. First, the decoder-only architecture inherently model long-range dependencies through autoregressive attention, enabling them to capture discourse-level cues and implicit argumentative structures that may span multiple sentences. This is particularly advantageous for ACD, where argumentative boundaries are often not signaled by explicit markers and may depend on broader contextual interpretation. Second, by casting the task as structured text generation with explicit XML-style tagging, the model learns to internalize both boundary detection and labeling decisions jointly, rather than propagating errors across separate pipeline stages. This reduces cumulative segmentation–classification error and encourages globally coherent predictions.

\subsection{Qualitative study}
A manual inspection of model outputs revealed several noteworthy qualitative phenomena that are not fully captured by standard automatic metrics. Beyond typical correct and incorrect label assignments (Table \ref{tab:errors-qualit-results}), we observed three recurring patterns that highlight both the strengths and limitations of the generative formulation.

The first phenomenon concerns what we refer to as argument type refinement. In several cases, the model predicts a component label that differs from the gold annotation but is arguably more appropriate given the context. In such instances, the generated label exposes potential inconsistencies or borderline cases in the original annotation. An example is illustrated in Tables \ref{tab:type-enhancement-and-rephrasing-qualit-results} and \ref{tab:type-enhancement-1-qualit-results}. Although these predictions are currently penalized as errors under strict span-level evaluation, they may reflect legitimate interpretative alternatives rather than genuine mis-labelling. This observation suggests that instruction-tuned LLMs can implicitly learn nuanced distinctions in argumentative structure and may even help identify annotation noise or ambiguities in existing corpora.

\begin{table}[ht]
    \centering
    \begin{tabular}{|p{0.1\textwidth}p{0.34\textwidth}|}
    \hline
        Gold &  And so \textcolor{cyan}{<premise>}if you believe the same thing\textcolor{cyan}{</premise>}, \textcolor{red}{<claim>}you just don't want to raise taxes on people\textcolor{red}{</claim>}. And \textcolor{red}{<claim>}the reality is it's not just wealthy people\textcolor{red}{</claim>}\\\hline
        OPT-6.7B ACD & And so \textcolor{cyan}{<premise>}if you believe the same thing\textcolor{cyan}{</premise>}, \textcolor{red}{<claim>}you just don't want to raise\textcolor{red}{</claim>} taxes on people. And \textcolor{cyan}{<premise>}the reality is it's not just wealthy people\textcolor{cyan}{</premise>}\\\hline
    \end{tabular}
    \caption{Example of ACD with argument type errors}
    \label{tab:errors-qualit-results}
\end{table}

\begin{table}[ht]
    \centering
    \begin{tabular}{|p{0.1\textwidth}p{0.34\textwidth}|}
    \hline
        Gold & \textcolor{red}{<claim>}We will do what we do best</claim>. \textcolor{red}{<claim>}It's a strategy that we've been working on for a \textbf{couple of years}\textcolor{red}{</claim>}. \textcolor{red}{<claim>}It is going to take us to much better advantage in conventional forces\textcolor{red}{</claim>}\\\hline
        OPT-6.7B ACD & \textcolor{red}{<claim>}We will do what we did best\textcolor{red}{</claim>}. \textcolor{cyan}{<premise>}It's a strategy that we've been working on for a \textbf{few years}\textcolor{cyan}{</premise>}. \textcolor{red}{<claim>}It is going to take us to much better advantage in conventional forces\textcolor{red}{</claim>}\\\hline
    \end{tabular}
    \caption{Example of ACD with argument type refinement and lexical adjustment}
    \label{tab:type-enhancement-and-rephrasing-qualit-results}
\end{table}

The second phenomenon relates to surface-level textual changes. The models occasionally introduce minor grammatical or lexical adjustments during generation, such as correcting verb conjugations, modifying singular/plural forms, or slightly reformulating phrases (see Table \ref{tab:type-enhancement-and-rephrasing-qualit-results}). While these changes preserve the semantic content and argumentative role of the text, they cause exact string mismatches when compared token-by-token with the reference output. Because our current evaluation protocol relies on strict span alignment, such variations are counted as errors despite representing semantically equivalent predictions and is considered as one of the limitations of our approach.

The third phenomenon involves the identification of previously unannotated argumentative components. In some cases, the model marks spans as claims or premises that were not labeled in the ground-truth data but should reasonably be interpreted as argumentative. This behavior demonstrates the model’s capacity to generalize argumentative patterns beyond the annotated instances and to detect implicit argumentative structures that may have been overlooked during manual annotation. While these predictions are again treated as false positives under the current evaluation framework, they highlight the model’s ability to capture deeper discourse-level reasoning. Illustrative examples are shown in Tables \ref{tab:discovery-qualit-results} and \ref{tab:discovery-1-qualit-results}.

\begin{table}[ht]
    \centering
    \begin{tabular}{|p{0.1\textwidth}p{0.34\textwidth}|}
    \hline
        Gold & \textcolor{red}{<claim>}Maybe we need to do a better job in mental clinics to help them\textcolor{red}{</claim>}. Because \textcolor{cyan}{<premise>}there is a major problem there\textcolor{cyan}{</premise>}\\\hline
        
        OPT-6.7B ACD & \textcolor{red}{<claim>}Maybe we need to do better job in mental clinics\textcolor{red}{</claim>} \textbf{\textcolor{cyan}{<premise>}to help them\textcolor{cyan}{</premise>}}. Because \textcolor{cyan}{<premise>}there is a major problem there\textcolor{cyan}{</premise>}.\\\hline
    \end{tabular}
    \caption{Example of ACD with argument components discovery}
    \label{tab:discovery-qualit-results}
\end{table}

A final, less frequent yet a possible limitation of our approach concerns hallucination during generation. In some cases, despite explicit instructions to preserve the original input verbatim and only insert XML tags, the models introduce new words, modify existing tokens, or generate short additional sub-sequences. This behavior alters the surface form of the text and leads to span misalignment, thereby negatively affecting the evaluation scores. We experimented with several mitigation strategies, including lowering the temperature, restricting nucleus sampling, strengthening prompt constraints, and explicitly emphasizing verbatim reproduction in the instructions. However, these adjustments did not fully eliminate the issue. This suggests that the phenomenon is likely intrinsic to autoregressive generative models, which are optimized for fluent continuation rather than strict copying. An illustrative example of this limitation is provided in Table \ref{tab:hallucination-qualit-results}. While relatively rare, such hallucinations remain a key challenge for generative formulations of ACD and point to the need for future work on constrained decoding strategies or hybrid extractive–generative approaches that better enforce input fidelity.
\begin{table}[ht]
    \centering
    \begin{tabular}{|p{0.1\textwidth}p{0.34\textwidth}|}
    \hline
        Gold & \textcolor{cyan}{<premise>}She's been doing this for 30 years\textcolor{cyan}{</premise>} \\\hline
        OPT-6.7B ACD & \textcolor{cyan}{<premise>}She's been doing this \textbf{job} for 30 years\textcolor{cyan}{</premise>}\\\hline
    \end{tabular}
    \caption{Example of ACD with hallucination}
    \label{tab:hallucination-qualit-results}
\end{table}
\subsection{Class-wise ACD}
A more fine-grained analysis of component detection performance under the BIO tagging scheme is provided in Table \ref{tab:comp_det_res_detailed}, which enables a closer examination of boundary detection accuracy and class-specific performance.

On the PE dataset, instruction-tuned LLMs clearly outperform BERT-based baselines across all argument classes. In particular, Llama-3-8B achieves the strongest performance, with balanced scores for both claim boundaries (B-C: 0.84, I-C: 0.80) and premise boundaries (B-P: 0.90, I-P: 0.88), resulting in the highest macro F1 (0.88). Notably, premise components (B-P and I-P) are generally detected more accurately than claims across all generative models, suggesting that supporting arguments exhibit more consistent lexical or structural cues. The O class consistently obtains the highest scores (up to 0.98), which is expected given its higher frequency and clearer separation from argumentative spans. In contrast, DeBERTa-v3 and RoBERTa show substantially lower performance on B- and I- tags, especially for claim beginnings, indicating difficulty in precise span delimitation.
On the merged dataset, all models experience a moderate drop in performance, as discussed previously. However, generative LLMs maintain relatively stable and well-balanced performance across both boundary markers (B- and I- tags) and argumentative classes (claims and premises). Unlike results on the PE dataset, wher premises are consistently easier to detect than claims, the merged configuration shows a more uniform distribution of scores across component types, suggesting that the models learn a more generalized representation of argumentative structure.
\begin{table}[ht]
    \centering \scriptsize
    \resizebox{.485\textwidth}{!}{%
    \begin{tabular}{|c|c|c|c|c|c|c|c|}
    \hline
        Model & B-C & I-C & B-P & I-P & O & F1-Macro \\\hline
        OPT-6.7B (PE) & 0.78 & 0.76 & 0.87 & 0.88 & 0.97 & 0.85 \\\hline
        GPT-2-1.5B (PE) & 0.78 & 0.76 & 0.87 & 0.89 & 0.96 & 0.85 \\\hline
        Llama-3-8B (PE) & 0.84 & 0.80 & 0.90 & 0.88 & 0.98 & 0.88 \\\hline
        DeBERTa-v3 (PE)	& 0.6 & 0.74 & 0.58	& 0.79 &  0.82 & 0.71 \\\hline
        RoBERTa (PE) & 0.59	& 0.68	& 0.59 & 0.78 &  0.19 & 0.69 \\\hline\hline
        OPT-1.3B (Merge) & 0.75 & 0.73 & 0.71 & 0.73 & 0.92 & 0.77 \\\hline
        OPT-6.7B (Merge) & 0.77 & 0.73 & 0.74 & 0.74 & 0.93 & 0.78 \\\hline
        GPT-2-1.5B (Merge) & 0.76 & 0.73 & 0.72 & 0.72 & 0.92 & 0.77 \\\hline
        Llama-3-8B (Merge) & 0.73 & 0.68 & 0.76 & 0.76 & 0.91 & 0.77 \\\hline
        DeBERTa-v3 (PE) & 0.41 & 0.43 & 0.46 & 0.5 &    0.63 & 0.49 \\\hline
        RoBERTa (PE) & 0.40 & 0.40 & 0.46 & 0.49 & 0.66 & 0.48 \\\hline
        
    \end{tabular}}
    \caption{Detailed f1-scores for ACD by experiment configuration}
    \label{tab:comp_det_res_detailed}
\end{table}

\section{Conclusion}
In this work, we revisited Argumentative Component Detection (ACD) in Argumentation Mining through the lens of instruction-tuned large language models (LLMs). Departing from traditional sequence labeling and multi-stage pipeline approaches, we reformulated ACD as a unified generative task that jointly performs argument delimitation and classification directly from raw text. Our experimental results demonstrate that instruction-tuned open-weight LLMs can effectively handle this structured prediction task. Our best performing model achieved a macro-F1 score of 0.8778, surpassing all considered baselines and approaching the human gold standard. Our study also revealed additional strengths of the generative formulation where the models occasionally corrected annotation inconsistencies and identified plausible yet unannotated argumentative components. This work represents one of the early attempts to fully cast ACD as a generative task, opening promising directions for future research.

\section*{Limitations}
Despite the encouraging results in this work, several limitations must be acknowledged. Hallucination remains an inherent limitation of LLMs. In rare cases, models introduce new words or modify the original input despite explicit instructions to replicate the input text without any alteration. Such deviations disrupt span alignment and negatively impact the evaluation scores. While decoding constraints (low temperature, restricted top-p) helped us to reduce this behavior, this problem still need further research endeavours. Our experiments focus exclusively on claim and premise detection. While this choice was intentional to prioritize unified argument component extraction, other important argument mining subtasks such as relation identification, stance modeling, and full argument graph construction remain unexplored within the generative framework.

\section*{Ethical considerations}
The datasets used in this work might reflect domain/language-specificities, and argumentative annotations inherently involve subjective judgments. Consequently, the models may reproduce or amplify potential existing biases and should not be interpreted as providing objective or universally valid argument structures. Also, argument boundary identification is inherently interpretative, and generative models may expose annotation inconsistencies or implicitly reshape argumentative structure, potentially presenting subjective analyses as objective outputs. Moreover, as observed in our experiments, LLMs may hallucinate or slightly alter input text, which can distort meaning in sensitive contexts. We emphasize that the proposed system is intended for analytical and research purposes, as argumentation technologies may be misused in contexts such as political persuasion or large-scale rhetorical analysis, particularly when combined with generative capabilities.

\bibliography{bibliography}
\newpage
\section*{Appendix}
\begin{table}[ht]
\centering
\resizebox{.485\textwidth}{!}{%
\begin{tabular}{|l|l|l|l|l|l|}
\hline
    Dataset & O & B-Premise & I-Premise & B-Claim & I-Claim\\\hline
    USElecDeb60To16 & 566492 & 26055 & 350079 & 29624 & 338941 \\ \hline
    Persuasive Essays & 35946 & 2257 & 29828 & 3832 & 59652 \\ \hline
    Web Discourse & 61414 & 195 & 3491 & 538 & 20566 \\ \hline 
\end{tabular}}
\caption{Detailed statistics on the datasets used for fine-tuning and testing LLM models following the BIO-tagging scheme}
\label{tab:comp_data_detailed}
\end{table}

\begin{figure*}[ht]
    \centering
    \includegraphics[width=\linewidth]{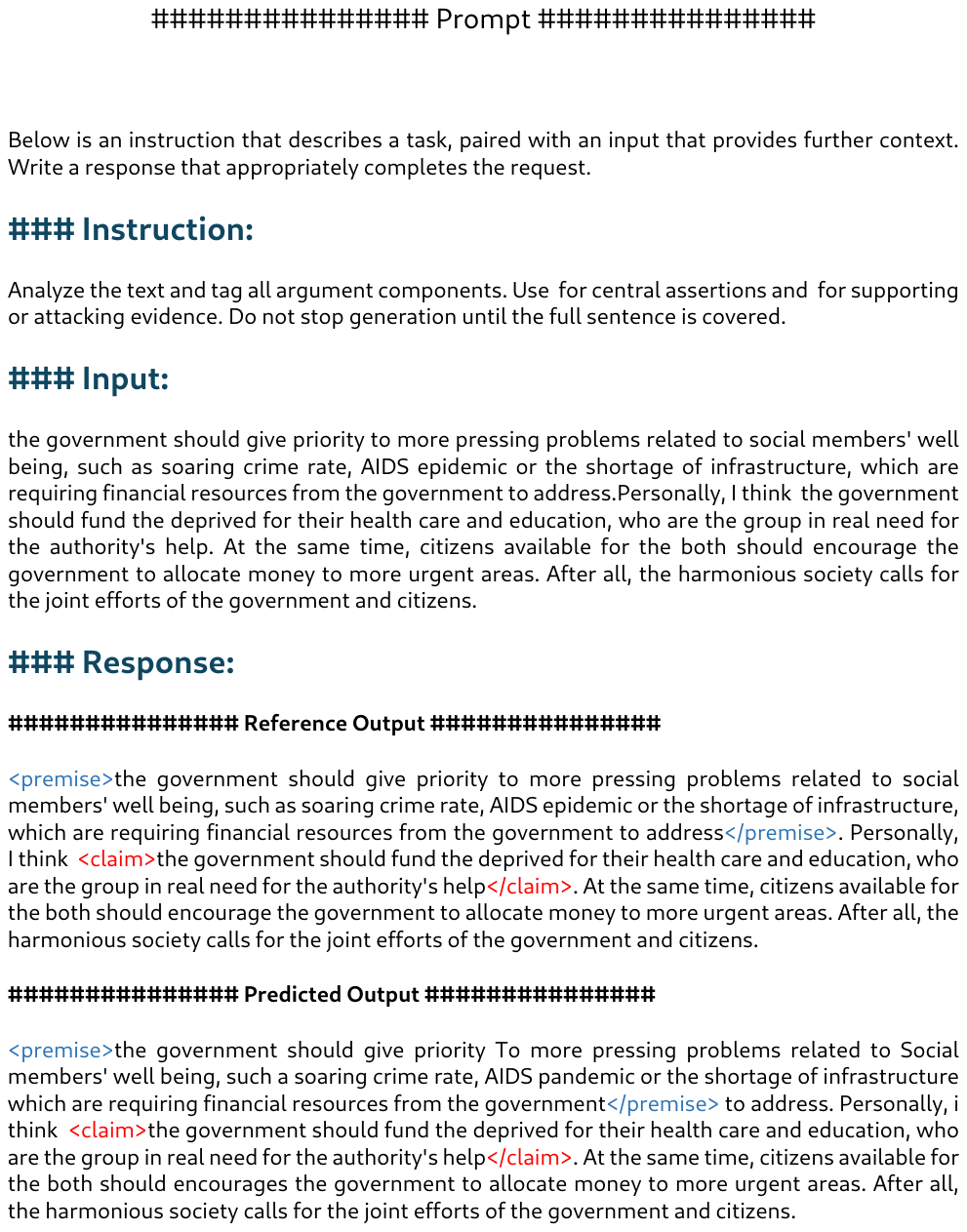}
    \caption{Full prompt and generation example for ACD task in AM}
    \label{fig:eg_prompt_answer}
\end{figure*} 

\begin{table}[ht]
    \centering
    \begin{tabular}{|p{0.1\textwidth}p{0.34\textwidth}|}
    \hline
        Gold & As a former educator, I believe that every child deserves equal access to education in the least restrictive environment possible. As a parent, I support that concept as long as the inclusion doesn't come at the expense of the other students in class. Inclusion can be successful rewarding for all involved but it also has drawbacks.\\\hline
        
        OPT-6.7B ACD & \textcolor{red}{<claim>}As a former educator,\textcolor{red}{</claim>} \textcolor{red}{<claim>}I believe that every child deserves equal access to education in the less restrictive environment possible.\textcolor{red}{</claim>}\textcolor{red}{<claim>}As a parent,\textcolor{red}{</claim>} \textcolor{cyan}{<premise>}I support that concept as long as the inclusion doesn't come at the expense of the other children in class.\textcolor{cyan}{</premise>}\textcolor{red}{<claim>}Inclusion can be successful rewarding for all involved but also has drawbacks.\textcolor{red}{</claim>}\\\hline
    \end{tabular}
    \caption{A second example of ACD with argument components discovery}
    \label{tab:discovery-1-qualit-results}
\end{table}

\begin{table}[ht]
    \centering
    \begin{tabular}{|p{0.1\textwidth}p{0.34\textwidth}|}
    \hline
        Gold & \textcolor{cyan}{<premise>}I'm not a politician\textcolor{cyan}{</premise>}, but \textcolor{red}{<claim>}I think I could go to Washington in a week and get everybody holding hands and get this bill signed because I talk to the Democratic leaders and they want it\textcolor{red}{</claim>}\\\hline
        OPT-6.7B ACD & \textcolor{cyan}{<premise>}I'm not a politician\textcolor{cyan}{</premise>}, but \textcolor{red}{<claim>}I think I could go to Washington in a week and get Everybody holding hands and get this Bill signed\textcolor{red}{</claim>} \textcolor{cyan}{<premise>}because I talk to the Democratic Leaders and they want it\textcolor{cyan}{</premise>}\\\hline
    \end{tabular}
    \caption{A second example of ACD with argument type refinement}
    \label{tab:type-enhancement-1-qualit-results}
\end{table}
\end{document}